\pdfoutput=1

\documentclass[11pt]{article}

\usepackage{acl}

\usepackage{microtype}
\usepackage{graphicx}
\usepackage{subfigure}
\usepackage{booktabs} 
\usepackage{times}
\usepackage{latexsym}
\usepackage{enumitem}
\usepackage{xcolor,colortbl}
\usepackage[T1]{fontenc}

\usepackage[utf8]{inputenc}

\usepackage[textsize=tiny]{todonotes}
\usepackage{amsmath}
\usepackage{amssymb}
\usepackage{mathtools}
\usepackage{amsthm}
\usepackage{multirow}  
\theoremstyle{plain}

\theoremstyle{definition}

\theoremstyle{remark}

\newcommand{\method}[0]{ALSACE~}
\newcommand{\model}[0]{ALSACE~}

\newcommand{\graycell}[1]{\cellcolor{gray!15}#1}
\title{Mitigating Language-Level Performance Disparity in mPLMs via Teacher Language Selection and Cross-lingual Self-Distillation}

\author{
Haozhe Zhao$^\ast$$^{1,2}$, 
Zefan Cai$^\ast$$^{1,2}$, 
Shuzheng Si$^\ast$$^{1,2}$,
Liang Chen$^1$, \\
\textbf{Yufeng He}$^{1,2}$,
\textbf{Kaikai An}$^{1,2}$,
\textbf{Baobao Chang}$^\dagger$$^{1,3}$
\\
$^1$National Key Laboratory for Multimedia Information Processing, Peking University \\
$^2$School of Software and Microelectronics, Peking University, China \\
$^3$Collaborative Innovation Center for Language Ability, Xuzhou, 221009, China \\
\texttt{\{mimazhe55360,zefncai\}@gmail.com}, 
\texttt{sishuzheng@stu.pku.edu.cn} \\
\texttt{chbb@pku.edu.cn}
}

\begin{document}
\maketitle
\renewcommand{\thefootnote}{\fnsymbol{footnote}}
\footnotetext[1]{Equal contribution.}
\footnotetext[2]{Corresponding author.}
\renewcommand{\thefootnote}{\arabic{footnote}}

\begin{abstract}
 
Large-scale multilingual Pretrained Language Models (mPLMs) yield impressive performance on cross-language tasks, yet significant performance disparities exist across different languages within the same mPLM. 
Previous studies endeavored to narrow these disparities by supervise fine-tuning the mPLMs with multilingual data.
However, obtaining labeled multilingual data is time-consuming, and fine-tuning mPLM with limited labeled multilingual data merely encapsulates the knowledge specific to the labeled data.
Therefore, we introduce \textbf{ALSACE} to leverage the learned knowledge from the well-performing languages to guide under-performing ones within the same mPLM, eliminating the need for additional labeled multilingual data. 
Experiments show that ALSACE effectively mitigates language-level performance disparity across various mPLMs while showing the competitive performance on different multilingual NLU tasks, ranging from full resource to limited resource settings. 
The code for our approach is available at \url{https://github.com/pkunlp-icler/ALSACE}.

\end{abstract}
\section{Introduction}


Recently, Multilingual Pre-trained Language Models (mPLMs) have attracted significant attention ~\citep{doddapaneni2021primer}.
These mPLMs, such as  mBERT~\citep{devlin2018bert} and mT5~\cite{xue2020mt5}, are pre-trained on extensive amounts of corpus across hundreds of different languages, which enables them to handle multiple languages within a single model and effectively perform cross-lingual tasks~\citep{lewis2019mlqa,zhang2020improving,stickland2020recipes,mutuvi2020multilingual,brown2020language,choudhury2021linguistically}.


However, all mPLMs share a key limitation. Due to discrepancies in the quality and quantity of pre-training corpus available for different languages, there is a noticeable performance disparity among different languages for the same mPLM, especially when comparing the performance of high-resource languages to that of low-resource languages.
For example, in Cross-lingual Natural Language Inference (XNLI) task~\citep{conneau2018xnli}, 
high-resource languages such as English can achieve a performance advantage of approximately 15 points compared to low-resource languages like Swahili, even within the same mPLM.



Several works have been proposed to investigate the reason for the performance disparity. \citet{kassner2019negated,wallat2021bertnesia,kassner2021multilingual} demonstrate that mPLMs could learn language-specific knowledge from different languages' pre-training corpus, but the imbalance of the corpus for different languages leads to the knowledge disparity for different languages. 
Therefore, ~\citet{kassner2021multilingual} suggests the observed language-level performance disparity can be attributed to the disparity of learned different languages knowledge during the pre-training stage.
Therefore, \citet{dong-etal-2021-data,hu-etal-2021-explicit} attempts to narrow the knowledge disparity by involving additional supervised data in different languages to fine-tune the mPLM. 
However, obtaining such labeled multilingual data is time-consuming and expensive. 
Moreover, these labeled data mostly come from limited tasks and domains, which makes it hard to compensate for the large knowledge disparity during the pre-training stage, restricting the generalization performance of the low-resource languages on downstream tasks.
\begin{figure*}[htbp]
    \centering
    \includegraphics[scale=0.45]{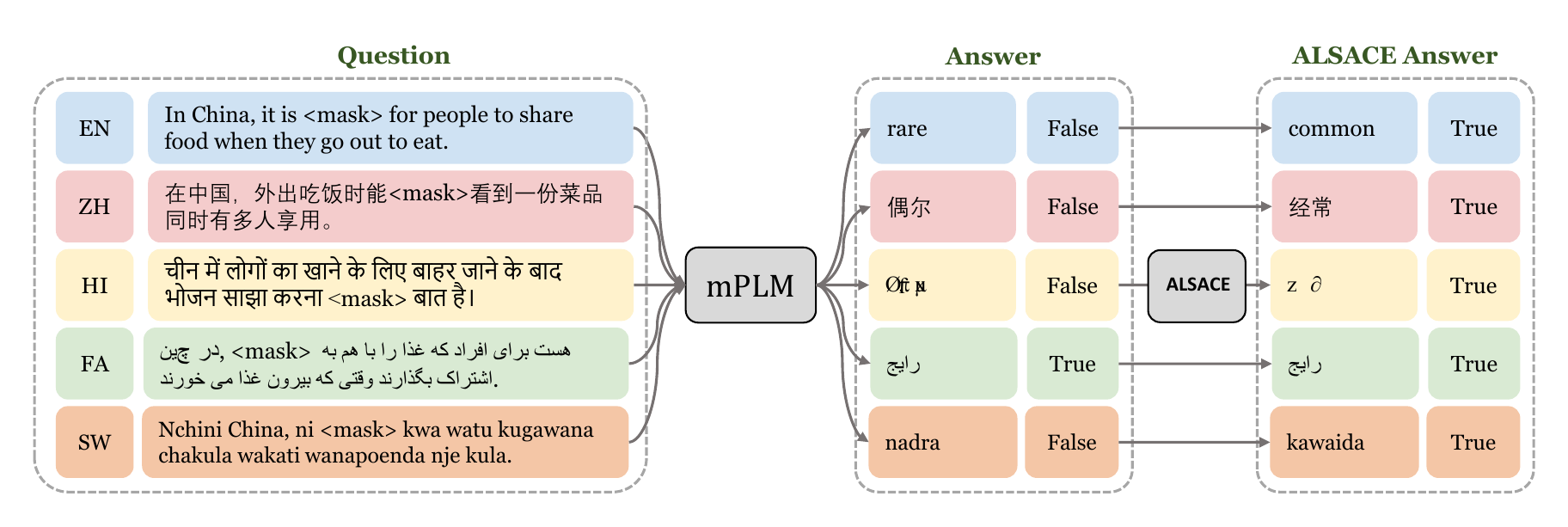}
    \caption{
     ~\method can reduce language-level performance disparity via mitigating knowledge disparity across languages on GeoMLAMA benchmark ~\citep{yin2022geomlama}.
    }\label{figure:method}
\end{figure*}
To utilize the different knowledge across different languages within the same mPLM and mitigate the need for the labeled data, we introduce te\textbf{A}cher \textbf{L}anguage \textbf{S}election \textbf{A}nd \textbf{C}ross-lingual s\textbf{E}lf-distillation (\textbf{ALSACE}), which leverages the knowledge from the selected teacher languages to reduce the performance disparity among the languages.
Specifically, ALSACE mainly consists of two stages: Teacher Language Selection and Cross-Lingual Self-Distillation.
For teacher language selection, the motivation is that high-resource languages may not be ideal for probing knowledge to supervise the other languages. 
For instance, although Persian is a relatively low-resource language, it may provide more precise answers for Kenya's cultural queries than English due to the closer linguistic proximity ~\citep{yin2022geomlama} between Persian and Swahili.
Different from simply using the knowledge from high-resource languages (e.g., English) to improve the performance of low-resource languages (e.g., Swahili), we introduce Teacher Language Selection to choose reliable teacher languages for a specific task to supervise the student languages.
Specifically, we employ a majority voting strategy to generate pseudo-labels derived from the consensus of the mPLMs' predictions across different languages in the given multilingual corpus.
Then, we utilize the average confidence score of the different languages on the generated pseudo labels as the indicator to select the teacher languages automatically.
As a result, we can select adaptive teachers for different tasks using the unlabeled sentences in the corpus.
We further propose Cross-Lingual Self-Distillation to leverage the knowledge from each selected teacher language to supervise other languages, reducing the performance disparity.
We further propose cross-lingual self-distillation to leverage the knowledge from each selected teachers languages to supervise other languages, reducing the performance disparity.
It employs a consistency loss that encourages closer alignment between the model output distributions of each reliable teacher language and other languages. 
In this way, mPLMs can effectively mitigate the language-level performance disparity without relying on the supervised multilingual data.

Experiments show ALSACE consistently mitigates language-level performance disparity in various mPLMs and show the competitive performance on different multilingual benchmarks, including XNLI~\cite{conneau2018xnli}, PAWS-X~\cite{yang2019paws} and XCOPA~\citep{ponti-etal-2020-xcopa}.
We also conduct knowledge probing experiments on the GeoMLAMA~\cite{yin2022geomlama} as shown in Figure~\ref{figure:method}, demonstrating that ALSACE effectively mitigates language-level performance disparity by addressing knowledge disparity.
Moreover, our experiments show that ALSACE improves performance not only in low-resource languages but also in high-resource languages.
This finding illustrates that ALSACE enables effective knowledge transfer between different languages instead of only transferring knowledge from high-resource to low-resource languages.
Further analysis shows that ALSACE can transfer both general knowledge across different languages and language-specific knowledge, i.e., some specific knowledge locally shared by people speaking the specific language, which is only present in the corpus of some specific languages.
\section{Related Work}
\label{sec:related_work}

\textbf{Knowledge Disparity Leads to Language-Level Performance Disparity in mPLMs.}
The mPLMs have shown strong capabilities in many NLP tasks including Natural Language Generation (NLG) \citep{si-etal-2022-mining, si2024spokenwoz, zhao2023mmicl, cai2023unipcm,song23c_interspeech, li2024shot,liu2023mlbench} and natural language understanding (NLU) \citep{si-etal-2022-scl,si2023santa,liu2023towards,an-etal-2023-coarse,hu2023distantlysupervised}. 
However, there is a noticeable performance disparity across different languages in the same mPLM.
Several works are proposed to investigate the reason of language-level performance disparity in mPLMs.
\citet{wallat2021bertnesia,kassner2021multilingual} demonstrate that mPLMs could learn different knowledge from different languages data in the pre-training corpus, but imbalanced corpus might lead to knowledge disparity for different languages. 
~\citet{kassner2021multilingual} suggests that the performance disparities across different languages could be attributed to the imbalanced knowledge distribution of these languages acquired during the pre-training phase.
\citet{yin2022geomlama} further observe that different languages within a single mPLM can retain distinct knowledge that is locally shared by the people speaking the specific language.
Therefore, we attempt to address language-level performance disparity from the knowledge disparity perspective.



\noindent
\paragraph{Mitigating Language-Level Performance Disparity in mPLMs.}
Previous studies have utilized cross-lingual knowledge to mitigate the language-level performance disparity.
 \citet{he-etal-2021-effectiveness} employ lightweight adapters on the mPLMs to mitigate forgetting issues.
{InfoXLM~\citep{chi2021infoxlm}} designs a new pre-training task with 42GB parallel data to align the representation of multiple languages.
{XLE~\citep{chi2022xlme}} pre-trains mPLMs with a generator and discriminator structure on 142B tokens. 
 These methods attempt to incorporate multilingual resources to mitigate performance disparity. 
 However, obtaining multilingual data can be time-consuming and restricts model performance on low-resource languages.
Thus,
~\citet{yang2022crosslingual,nguyen2021improving} attempt to enhance mPLMs by distilling knowledge from well-learned monolingual teachers.
~\citet{qi-etal-2022-enhancing} learn from different cross-lingual templates using consistency loss to enforce correspondence representation among languages.
Different from distilling knowledge from other monolingual models, we aim to reduce the language-level performance disparity within mPLMs.
\section{Method}
\label{section:zero_label}





\subsection{Teacher Language Selection}
\label{subsection:ensemble_based_score}

To mitigate the language-level performance disparity within mPLMs, we utilize knowledge from the appropriate teacher language to supervise other languages. 
An intuitive idea is to transfer the knowledge from high-resourced to low-resourced languages to mitigate the disparity.
However, due to the different linguistic proximity between different languages, the high-resource languages may not be ideal teachers for transferring knowledge to other languages in the specific task. 
For example, low-resourced Persian may provide more accurate responses to Kenya’s cultural queries compared to high-resource English, which makes it a better teacher language for Swahili than English.
Therefore, the proposed Teacher Language Selection aims to choose reliable teacher languages for a specific task to guide the student languages.

Considering the given corpus $D$ for the specific multilingual task (e.g., Cross-lingual Natural Language Inference) that spans over $T$ languages, we aim to utilize the proposed Teacher Language Selection to identify the teacher languages to mitigate language-level performance disparity efficiently.
Precisely, we first fine-tune the mPLMs with an English training set $D_{en}$ of the given task to obtain a better initialization.
We secondly utilize the mPLMs to generate the prediction $\hat{y}_{t,i}$ of the given instance $x_i$ from corpus $D$ in language $ t \in T$.
Then, we employ a majority vote strategy on the predictions of different languages to generate the pseudo label $y_{i}$ of the instance $x_{i} \in X$, as follows:
\begin{equation}  
\begin{aligned}  
\hat{y}_{t,i} &= \mathop{\mathrm{argmax}}_{y \in Y}P(y\,|\,x_{t,i}) \\  
y_{i} &= \underset{k}{\operatorname{argmax}} \sum_{t \in T} \mathbb{I}(\hat{y}_{t, i} = k)
\end{aligned}  
\end{equation}
where $P(y\,|\,x_{t,i})$ denotes the predicted probability of the given mPLM on instance $x_{t,i}$ in language $t$. $\mathbb{I}$ is the indicator function, while $k$ signifies the set of all possible results for the given task.  
The generated pseudo-labels reflect the collective understanding of the provided instance across various languages. 
Thus, it reduces the risk of incorrect pseudo-labeling compared to relying solely on the prediction from a single language (even a high-resource language like English).

We further employ the pseudo-labels to compute the average confidence score $s_t$ for each language, which allows us to assess the capabilities of different languages in the mPLM.
The average confidence score $s_t$ indicates the level of agreement between each language and the common understanding of the mPLMs, i.e., languages with a higher average confidence score are more likely to make accurate predictions for a given instance.
Ultimately, we normalize the confidence score and use the normalized score $\hat{s_t}$ to evaluate which languages demonstrate superior performance:
\begin{equation}  
\label{eq:theta}  
\begin{aligned}  
s_t &= \frac{1}{|X|}\sum_{x_{t,i}}^{X}P(y_{t,i}|x_{t,i} )  \\  
\hat{s_t}  &=  \frac{{e^{s_t}}}{{\sum_{j}^{T} e^{s_j}}}, \quad t \in T
\end{aligned}  
\end{equation}
where the $T$ refers to the collection of all languages involved in the given multilingual task. 
We set the threshold $\theta$ to be the average value of the normalized score $\hat{s_t}$ to select the teacher languages $T_\textit{teacher}$ and student languages $S_\textit{student}$, as follows:
\begin{equation}  
\label{eq:theta}  
\begin{aligned}  
T_\textit{teacher}  &=  \{ t | t \in T, \hat{s_t} \geq \theta\} \\  
T_\textit{student}  &=  \{ t | t \in T, \hat{s_t} < \theta\}  
\end{aligned}  
\end{equation}
In this way, we can automatically select appropriate teacher languages for the different multilingual tasks to mitigate language-level performance disparity efficiently.
Moreover, we do not need any labeled multilingual data to improve the cross-lingual transfer ability of mPLMs \citep{chi2022xlme,chi2021infoxlm}.







\subsection{Cross-Lingual Self-Distillation}
\label{subsection:zero_label_tuning}
Having selected the appropriate teacher languages for the given multilingual task, we further introduce Cross-Lingual Self-Distillation to leverage the knowledge from each selected teacher language to supervise other languages.
Specifically, we construct a parallel multilingual pair set $\hat{X}$ that consists of parallel sentence pairs between each two languages. 
To reduce the disturbance caused by student languages, we exclusively employ parallel pairs of teacher-student and teacher-teacher languages as potential candidates for self-distillation.
Therefore, the instance pair $\hat{X}$ can be defined as: 
\begin{equation}
\small
\hat{X}=\{\text{ }(x_{t1,i}, x_{t2,i})\text{ } |\text{ }t_1 \in T,\text{ }t_2 \in T_\textit{teacher} ,\text{ } x_i \in X\text{ } \}
\end{equation}
where $T_\textit{teacher}$ is the selected teacher languages. 
We filter out student-student language pairs to prevent student languages from learning from each other.

For the selected candidate instance pairs, we use Kullback-Leibler divergence as a consistency loss to encourage closer alignment between the prediction distributions of the reliable teacher language and the target language.
In this way, mPLMs can effectively transfer and distill the knowledge from the teacher language to the target language, mitigating the language-level performance disparity.
The final consistency loss $\mathcal{L} $ can be formulated as follows:
\begin{equation}
\begin{aligned}
\mathcal{L}  = \frac{1}{|\hat{X}|}\sum_{\hat{x}_{1},\hat{x}_{2}}^{\hat{X}} \text{KL}(\text{P}(\hat{x}_{1})\text{} ||\text{P}(\hat{x}_{2})) 
\end{aligned} 
\end{equation} 
where $\text{KL}(\text{P}||\text{Q})$ is the Kullback-Leibler divergence function. $P(\hat{x}_1)$ and $P(\hat{x}_2)$ are the prediction distributions of the given mPLM for the inputs $\hat{x}_1$ and $\hat{x}_2$ in different languages, respectively.

\section{Experiment}
\label{section:experiment}

\subsection{Experimental Details}
\label{subsubsection:experiment_details}
\textbf{Datasets.}
\begin{table}[ht]
\centering
\resizebox{1\linewidth}{!}{
\begin{tabular}{lccc}
\toprule
\textbf{Task} &\textbf{Dataset} &  \textbf{Lang.} & \textbf{Metric} \\ \midrule
Natural Language Inference&XNLI             &  15                 & Acc.            \\
Commonsense Reasoning& XCOPA            &  10                 & Acc.            \\
Paraphrase Identification& PAWS-X           &  7                  & Acc.            \\
 Commonsense Probing& GeoMLAMA         &5                  & Acc.            \\  \bottomrule
\end{tabular}
}
\caption{The tasks involved in experiments.}
\vspace{-0.2in}
\label{tab:multilingual_benchmarks}
\end{table}
As shown in Table \ref{tab:multilingual_benchmarks}, our experiments are conducted on various multilingual benchmarks: XNLI~\cite{conneau2018xnli}, PAWS-X~\cite{yang2019paws}, XCOPA~\citep{ponti-etal-2020-xcopa} and GeoMLAMA~\cite{yin2022geomlama}.\\
\textbf{Experimental Settings.} 
We follow the cross-lingual transfer setting as~\citet{lauscher2020zero}, first fine-tuning the model with an English training set and directly evaluating the model on multilingual test sets.
We apply \model to the fine-tuned model using unlabeled multilingual inputs $X$ from $T$ languages in order to address the language-level performance disparity across those languages.  
Specifically, We firstly use data generation methods, Supergen~\cite{meng2022generating}, which employ a language model to automatically generate text based on label-descriptive prompts, producing monolingual unlabeled data. 
Next, we use machine translation\footnote{The translation API from http://api.fanyi.baidu.com/ is utilized for generating multilingual parallel data.} to translate generated monolingual data and create unlabeled parallel multilingual pairs. 
By combining the data generation method and machine translation system, we establish an automated pipeline for generating unlabeled parallel corpora with minimal cost.\\
\textbf{Baselines.} We take the XLM-Align~\citep{chi2021improving},  $\text{XLMR-adapter}_{256}$~\citep{he-etal-2021-effectiveness}, InfoXLM~\citep{chi2021infoxlm}, VECO~\citep{luo2021veco}, ERNIE-M~\citep{ouyang2021erniem} 
and XLE~\citep{chi2022xlme} as baselines. 

Details can be found in Appendix~\ref{appendix:Details} and~\ref{appendix:baselines}.
\begin{table*}[h]
\centering
\small

\setlength\tabcolsep{3pt}
    \centering
    \resizebox{\linewidth}{!}{
        \begin{tabular}{lcccccccccccccccccc}
        \toprule
          Method & $ \text{Params}$ & Perf. &  en & fr & es & de & el & bg & ru & tr & ar & vi & th & zh & hi & sw & ur & \textbf{avg} \\ 
        \midrule
          
            \multirow{2}{*}{XLM-R-base}& \multirow{2}{*}{225M}& ~   & 84.23 & 77.39 & 78.20 & 76.45 & 75.97 & 77.80 & 75.35 & 73.27 & 71.84 & 74.93 & 71.88 & 74.23 & 69.22 & 64.55 & 65.77 & 74.07 \\
~ & ~ & \graycell{$\Delta \downarrow$} & \graycell{$\backslash$} & \graycell{6.84} & \graycell{6.03} & \graycell{7.78} & \graycell{8.26} & \graycell{6.43} & \graycell{8.88} & \graycell{10.96} & \graycell{12.39} & \graycell{9.30} & \graycell{12.35} & \graycell{10.00} & \graycell{15.01} & \graycell{19.68} & \graycell{18.46} & \graycell{10.88 (\textcolor{red}{-})} \\

            \multirow{2}{*}{XLM-Align} &\multirow{2}{*}{225M}& ~  & 86.70 & 80.60 & 81.00 & 78.80 & 77.40 & 78.80 & 77.40 & 75.20 & 73.90 & 76.90 & 73.80 & 77.00 & 71.90 & 67.10 & 66.60 & 76.30 \\ 
~ & ~ & \graycell{$\Delta \downarrow$} & \graycell{$\backslash$} & \graycell{\bf 6.10} & \graycell{\bf5.70} & \graycell{7.90} & \graycell{9.30} & \graycell{7.90} & \graycell{9.30} & \graycell{11.50} & \graycell{12.80} & \graycell{9.80} & \graycell{12.90} & \graycell{9.70} & \graycell{14.80} & \graycell{19.60} & \graycell{20.10} & \graycell{11.24 (\textcolor{red}{+0.36})} \\

           
            \multirow{2}{*}{\method-base} &\multirow{2}{*}{225M}& ~  &  84.11 & 77.80 & 78.30 & 77.50 & 76.51 & 78.28 & 76.01 & 74.19 & 72.12 & 75.50 & 72.81 & 74.90 & 70.12 & 65.55 & 66.37 & 74.67 \\ 
           
 ~ & ~ & \graycell{$\Delta \downarrow$} & \graycell{$\backslash$} & \graycell{\underline{6.31}} & \graycell{\underline{5.81}} & \graycell{\textbf{6.61}} & \graycell{\textbf{7.60}} & \graycell{\textbf{5.83}} & \graycell{\textbf{8.10}} & \graycell{\textbf{9.92}} & \graycell{\textbf{11.99}} & \graycell{\textbf{8.61}} & \graycell{\textbf{11.30}} & \graycell{\textbf{9.21}} & \graycell{\textbf{13.99}} & \graycell{\textbf{18.56}} & \graycell{\textbf{17.74}} & \graycell{\textbf{10.11 (\textcolor{green!70!black}{-0.77})}} \\
            \midrule
            \multirow{2}{*}{XLM-R-large} &\multirow{2}{*}{550M}& ~  &  86.45 & 80.90 & 81.84 & 81.22 & 79.36 & 80.74 & 78.78 & 77.23 & 77.03 & 77.82 & 75.53 & 77.82 & 74.55 & 69.62 & 70.86 & 77.98 \\ 
~& ~& \graycell{$\Delta \downarrow$} & \graycell{$\backslash$} & \graycell{5.45} & \graycell{4.51} & \graycell{5.13} & \graycell{6.99} & \graycell{5.61} & \graycell{7.57} & \graycell{9.12} & \graycell{9.32} & \graycell{8.53} & \graycell{10.82} & \graycell{8.53} & \graycell{11.80} & \graycell{16.73} & \graycell{15.49} & \graycell{8.97 (\textcolor{red}{-})}\\

     
            \multirow{2}{*}{XLM-R-adapter}  &\multirow{2}{*}{567M}& ~  &  89.22 & 83.27 & 84.69 & 83.47 & 82.39 & 83.59 & 79.74 & 78.80 & 78.62 & 79.32 & 77.84 & 79.34 & 76.42 & 72.21 & 72.27 & 80.08 \\ 
    
~ & ~ & \graycell{$\Delta \downarrow$} & \graycell{$\backslash$} & \graycell{5.95} & \graycell{4.53} & \graycell{5.75} & \graycell{6.83} & \graycell{5.63} & \graycell{9.48} & \graycell{10.42} & \graycell{10.60} & \graycell{9.90} & \graycell{11.38} & \graycell{9.88} & \graycell{12.80} & \graycell{17.01} & \graycell{16.95} & \graycell{9.79(\textcolor{red}{+0.82})} \\ 

            \multirow{2}{*}{Info-XLM-large} &\multirow{2}{*}{550M} & ~  &  89.70 & 84.50 & 85.50 & 84.10 & 83.40 & 84.20 & 81.30 & 80.90 & 80.40 & 80.80 & 78.90  & 80.90 & 77.90 & 74.80 & 73.70 & 81.40 \\ 
~ & ~ & \graycell{$\Delta \downarrow$} & \graycell{$\backslash$} & \graycell{5.20} & \graycell{4.20} & \graycell{5.60} & \graycell{6.30} & \graycell{5.50} & \graycell{8.40} & \graycell{8.80} & \graycell{9.30} & \graycell{8.90} & \graycell{10.80} & \graycell{8.80} & \graycell{11.80} & \graycell{14.90} & \graycell{16.00} & \graycell{8.89 (\textcolor{green!70!black}{-0.08})} \\

            \multirow{2}{*}{VECO-large}  &\multirow{2}{*}{550M}& ~  &  88.20 & 82.80	&84.20	&82.90	&81.20	&83.10&	80.30&	78.40&	79.20&	80.40&	77.00	&79.10	&76.20	&74.30	&71.30 & 79.90 \\ 
         
~ & ~ & \graycell{$ \Delta \downarrow$} & \graycell{$\backslash$} & \graycell{5.40}& \graycell{4.00}& \graycell{5.30}& \graycell{7.00}& \graycell{5.10}& \graycell{7.90}& \graycell{9.80}& \graycell{9.00}& \graycell{7.80}& \graycell{11.20}& \graycell{9.10}& \graycell{12.00}& \graycell{13.90}& \graycell{16.90} & \graycell{8.88(\textcolor{green!70!black}{-0.09})} \\ 

            \multirow{2}{*}{ERNIE-M-large}&\multirow{2}{*}{550M} & ~  &  89.30 & 85.10 & 85.70 & 84.40 & 83.70 & 84.50 & 82.00 & 81.20 & 81.20 & 81.90 & 79.20 & 81.00 & 78.60 & 76.20 & 75.40 & 82.00 \\ 

~ & ~ & \graycell{$\Delta \downarrow$} & \graycell{$\backslash$} & \graycell{4.20} & \graycell{3.60} & \graycell{4.90} & \graycell{5.60} & \graycell{4.80} & \graycell{7.30} & \graycell{8.10} & \graycell{8.10} & \graycell{7.40} & \graycell{10.10} & \graycell{8.30} & \graycell{10.70} & \graycell{\textbf{13.10}} & \graycell{13.90} & \graycell{7.86 (\textcolor{green!70!black}{-1.11})} \\
            
 
            \multirow{2}{*}{\method-large} &\multirow{2}{*}{550M}& ~  &  86.65 & 82.61 & 83.21 & 82.16 & 81.34 & 83.09 & 80.98 & 79.50 & 79.60 & 79.98 & 78.18 & 79.74 & 77.13 & 72.71 & 73.58 & 80.03 \\ 
~ & ~ & \graycell{$\Delta \downarrow$} & \graycell{$\backslash$} & \graycell{\textbf{4.04}} & \graycell{\textbf{3.44}} & \graycell{\textbf{4.49}} & \graycell{\textbf{5.31}} & \graycell{\textbf{3.56}} & \graycell{\textbf{5.67}} & \graycell{\textbf{7.15}} & \graycell{\textbf{7.05}} & \graycell{\textbf{6.67}} & \graycell{\textbf{8.47}} & \graycell{\textbf{6.91}} & \graycell{\textbf{9.52}} & \graycell{\underline{13.94}} & \graycell{\textbf{13.07}} & \graycell{\textbf{7.09 (\textcolor{green!70!black}{-1.88})}} \\

            \midrule
           \multirow{2}{*}{mT5-large}& \multirow{2}{*}{1.2B}& ~  &  88.42 & 82.44 & 83.49 & 81.68 & 81.14 & 81.96 & 79.90 & 77.33 & 76.87 & 78.52 & 75.31 & 77.74 & 75.31 & 72.63 & 70.88 & 78.91 \\ 

~ & ~ & \graycell{$\Delta \downarrow$} & \graycell{$\backslash$} & \graycell{5.98} & \graycell{4.93} & \graycell{6.74} & \graycell{7.28} & \graycell{6.46} & \graycell{8.52} & \graycell{11.09} & \graycell{11.55} & \graycell{9.90} & \graycell{13.11} & \graycell{10.68} & \graycell{13.11} & \graycell{15.79} & \graycell{17.54} & \graycell{10.19 (\textcolor{red}{-})}\\ 

     
            \multirow{2}{*}{XLE-large} &\multirow{2}{*}{840M}& ~  &  89.40 & 84.70 & 85.50 & 84.40 & 83.50 & 84.10 & 81.90 & 81.30 & 80.70 & 81.20 & 79.20 & 81.50 & 76.50 & 74.10 & 72.40 & 81.30 \\ 

~ & ~ & \graycell{$\Delta \downarrow$} & \graycell{$\backslash$} & \graycell{\textbf{4.70}} & \graycell{3.90} & \graycell{\textbf{5.00}} & \graycell{5.90} & \graycell{5.30} & \graycell{7.50} & \graycell{\textbf{8.10}} & \graycell{\textbf{8.70}} & \graycell{\textbf{8.20}} & \graycell{\textbf{10.20}} & \graycell{7.90} & \graycell{12.90} & \graycell{15.30} & \graycell{17.00} & \graycell{8.61 (\textcolor{green!70!black}{-1.58})} \\

            \multirow{2}{*}{\method-mT5} &\multirow{2}{*}{1.2B}& ~  &  88.60 & 83.69 & 84.79 & 83.17 & 82.91 & 83.91 & 81.80 & 79.54 & 78.84 & 80.20 & 77.90 & 80.92 & 77.25 & 75.17 & 73.13 & 80.79 \\ 
~ & ~ & \graycell{$ \Delta \downarrow$} & \graycell{$\backslash$} & \graycell{\underline{4.91}} & \graycell{\textbf{3.81}} & \graycell{\underline{5.43}} & \graycell{\textbf{5.69}} & \graycell{\textbf{4.69}} & \graycell{\textbf{6.80}} & \graycell{\underline{9.06}} & \graycell{\underline{9.76}} & \graycell{\underline{8.40}} & \graycell{\underline{10.70}} & \graycell{\textbf{7.68}} & \graycell{\textbf{11.35}} & \graycell{\textbf{13.43}} & \graycell{\textbf{15.47}} & \graycell{\textbf{8.37 (\textcolor{green!70!black}{-1.82})}} \\
        \bottomrule
        \end{tabular}
    }
    \caption{Main result of XLM-R-base, XLM-R-large, and mT5-large on XNLI dataset evaluated in accuracy. $\Delta$ represents the cross-lingual transfer gaps~\citep{chi2021infoxlm}, i.e., performance drop between English and other languages in zero-shot transfer. A smaller gap indicates better cross-lingual transferability. \method achieves competitive results compared to the state-of-the-art methods and enhances the performance of most languages across all three mPLMs, simultaneously reducing the language-level performance disparity amongst all the languages.}

    \label{table:main_results_xnli}
    
\end{table*} 

\subsection{Main Results}
\label{subsubsection:main_results}
\textbf{Overall Performance.}
\label{general_performance}
The results presented in Table~\ref{table:main_results_xnli} demonstrate that \method achieves the lowest cross-lingual transfer gaps across different baselines on XNLI for various mPLMs. 
\method yields an improvement of up to 0.6 points, 2.05 points, and 1.88 points, respectively, in average accuracy compared with XLM-R-base, XLM-R-large, and mT5-large baselines. 
Importantly, we achieve competitive performance with state-of-the-art methods across different mPLMs while improving the cross-lingual transferability of mPLMs without introducing any extra information.

For example, InfoXLM~\citep{chi2021infoxlm}, which is also based on XLM-R, uses 42GB of multilingual parallel data for pretraining. In contrast, \model depends solely on a small volume of unlabeled parallel data (500-shot), which can be automatically generated with minimal effort and and exhibits superior cross-lingual transferability compared to other baselines. 
While we also utilize parallel data to enhance cross-lingual transferability, our motivation diverges:
Instead of aligning multilingual representations through parallel data, our goal is to leverage the knowledge from teacher languages within mPLMs to supervise others.
The 500-shot unlabeled parallel data in \method are exclusively used to distill the knowledge of other languages in mPLMs. 
As a result, Table~\ref{table:main_results_xnli} shows performance enhancement and cross-lingual transfer gap reduction for most languages across different models. 
In comparison to state-of-the-art methods, \method does not mandate an extensive pre-training process or a large number of parallel corpora while achieving competitive performance and minimizing the cross-lingual transfer gaps. 
\begin{table*}[h]
\centering
\small
\begin{tabular}{lrr|rr|rr}
\toprule
\textbf{Method} & \textbf{avg(S)} $\uparrow$ & \textbf{$\Delta$(S) $\downarrow$} & \textbf{avg(T)}$\uparrow$ & \textbf{$\Delta$(T) $\downarrow$} & \textbf{avg(A)}$\uparrow$ & \textbf{$\Delta$(A) $\downarrow$} \\
\midrule
Excluding Weak in Stu. & 76.70 & 10.08 & 82.56 & 4.93 & 79.44 & 7.35 \\
Excluding Weak in Tea. & 76.92 & 9.73 & 82.50 & 4.84 & 79.52 & 7.13 \\
Random selection & 76.93 & 9.87 & 82.67 & 4.83 & 79.61 & 7.20 \\
No Selection & 77.05 & 9.83 & 82.69 & 4.90 & 79.69 & 7.20 \\
Scale-Based Selection & 77.13 & 9.80 & 82.73 & 4.89 & 79.75 & 7.18 \\
\rowcolor{gray!15} \model &\bf 77.55 & \bf9.10 & \bf 82.86 & \bf4.42 & \bf80.03 & \bf6.62 \\
\bottomrule
\end{tabular}
\caption{Ablation Study of the Teacher Language Selection. $\Delta$ represents the cross-lingual transfer gaps, i.e., performance drop between English and other languages in zero-shot transfer. A smaller gap indicates better cross-lingual transferability. We report the average performance and cross-lingual transfer gaps of the student languages(\textbf{S}), teacher languages(\textbf{T}), and all languages(\textbf{A}), respectively.}
\label{tab:ablation_of_teach}
\end{table*}
\begin{table*}[h]
\centering
\small
\begin{tabular}{lcc|cc|cc}
\toprule
\textbf{Method} & \textbf{avg(S)} $\uparrow$ & \textbf{$\Delta$(S) $\downarrow$} & \textbf{avg(T)}$\uparrow$ & \textbf{$\Delta$(T) $\downarrow$} & \textbf{avg(A)}$\uparrow$ & \textbf{$\Delta$(A) $\downarrow$} \\
\midrule
XLM-R-base      & 70.71 & 13.52 & 77.91 & 7.37  & 74.07 & 10.88 \\
E. Self-Train.  & 70.94 & 13.15 & 78.16 & 6.92  & 74.31 & 10.48 \\
F. Self-Train.  & 71.10 & 13.03 & 78.27 & 6.84  & 74.45 & 10.37  \\
\rowcolor{gray!15} ALSACE & \bf 71.44 & \bf12.67 & \bf78.35 & \bf6.72  & \bf74.67 & \bf10.12  \\
\midrule
XLM-R-large     & 75.06 & 11.39 & 81.33 & 5.98  & 77.98 & 9.07 \\
E. Self-Train.  & 75.82 & 10.95 & 81.74 & 5.87  & 78.58 & 8.77 \\
F. Self-Train.  & 75.89 & 10.92 & 82.10 & 5.50  & 78.79 & 8.60 \\
\rowcolor{gray!15} ALSACE & \bf 77.55 &\bf 9.10  & \bf82.86 & \bf4.42  & \bf80.03 & \bf7.09 \\
\midrule
mT5-large       & 75.57 & 12.85 & 82.72 & 6.65  & 78.91 & 10.19  \\
E. Self-Train.  & 76.55 & 11.95 & 83.21 & 6.18  & 79.66 & 9.48  \\
F. Self-Train.  & 76.81 & 11.83 & 83.32 & 6.21  & 79.85 & 9.42  \\
\rowcolor{gray!15} ALSACE & \bf 77.87 & \bf10.73 & \bf84.12 & \bf5.22  & \bf80.79 & \bf8.37  \\
\bottomrule
\end{tabular}
\caption{Comparison of self-distillation baselines with \model. 
$\Delta$ represents the cross-lingual transfer gaps.\\ \textbf{S}, \textbf{T}, and \textbf{A} stand for the set of student languages, teacher languages, and all languages, respectively.}
\label{table:self-distillation-ablation}
\end{table*}
\\
\newline
\textbf{Mitigating Languages-Level Performance Disparity.} \model effectively mitigates the language-level performance disparity of mPLMs and shows consistent improvements across different mPLMs in both high-resource and low-resource languages.
Specifically, not only do the student languages achieve higher-than-average improvements, but teacher languages also benefit from the guidance of their peers. 
Through self-distillation, \model facilitates cross-language knowledge transferring among both teacher and student languages. It also enables teacher languages to learn from each other. 
Even high-resource languages like French and Spanish have shown improvement across various mPLMs, which further supports this claim.
Notably, low-resource languages such as Swahili and Urdu experience substantial gains with \model, achieving improvements of 2.7 points and 2.4 points, respectively. These gains are particularly significant considering the relatively limited knowledge stored in multilingual pretrained language models (mPLMs) for these languages compared to other languages.  

Compared with other baselines, \model effectively reduces language-level performance disparities in mPLMs across various languages and minimizes the cross-lingual transfer gap. 
While some methods have enhanced overall performance, they have exacerbated the performance discrepancies between languages. 
They incorporated additional knowledge from the extensive parallel multilingual corpora into mPLMs. However, knowledge disparities persist and may even worsen, leading to increased cross-lingual transfer gaps.
We also perform \model across different tasks, such as PAWS-X and XCOPA. The result in Table~\ref{table:fewshot_results_pawsx} and Table~\ref{table:xcopa} shows that \model reduces the languages-level performance disparity of mPLMs.



\subsection{Ablation Study} 

\begin{table*}[h]
\centering
\small
\setlength\tabcolsep{3pt}
    \centering
        \scalebox{0.9}{
        \begin{tabular}{lcccccccccccccccc}
    \toprule
    Method & de & fr & ar & ru & zh & sw & en & vi & el & tr & bg & th & hi & ur & es & avg. \\
    \midrule
    XLM-R-large. & 69.84 & 69.52 & 65.18 & 69.00 & 66.43 & 61.65 & 71.81 & 67.75 & 68.76 & 66.27 & 70.28 & 63.17 & 63.94 & 61.85 & 69.24 & 66.98 \\
    E. Self-Train. & 69.06 & 68.48 & 65.67 & 68.32 & 65.85 & 60.10 & 71.86 & 67.96 & 68.40 & 65.39 & 69.40 & 62.46 & 64.35 & 61.96 & 69.02 & 66.55\\
    F. Self-Train. & 69.04 & 68.46 & 65.69 & 68.30 & 65.83 & 60.10 & 71.82 & 67.98 & 68.38 & 65.37 & 69.40 & 62.50 & 64.33 & 61.94 & 69.00 & 66.54 \\
    \rowcolor{gray!15} \textbf{\method} & \textbf{70.56} & \textbf{69.88} & \textbf{67.63} & \textbf{70.00} & \textbf{67.99} & \textbf{63.73} & \textbf{72.17} & \textbf{68.96} & \textbf{69.60} & \textbf{68.03} & \textbf{70.68} & \textbf{64.06} & \textbf{64.86} & \textbf{63.53} & \textbf{71.04} & \textbf{68.18} 
    \\
    \bottomrule
    \end{tabular}
    }
    \caption{
    \method performance on XLM-R-large in XNLI dataset under Limited-Resource Scenario. The metric in this table is accuracy. For each setting, we report the median scores among 5 runs.}
    \label{table:fewshot_results_xnli}
\end{table*} 
\begin{table*}[h]
\centering
\small
\setlength\tabcolsep{3pt}
    \centering
        \begin{tabular}{lcccccccc}
    \toprule
    Method & de & fr & zh & en & ko & ja & es & avg. \\
    \midrule
    XLM-R-large. & 82.35 & 82.75 & 77.05 & 85.60 & 73.24 & 72.70 & 83.40 & 79.58 \\
    E. Self-Train. & 82.35 & 82.50 & \textbf{78.05} & 85.95 & 72.44 & 72.80 & \textbf{83.70} & 79.68\\
    F. Self-Train. & 82.35 & \textbf{82.85} & 77.20 & 86.15 & \textbf{73.74} & 72.85 & 83.40 & 79.79 \\
    \rowcolor{gray!15} \textbf{\method} & \textbf{82.40} & \underline{82.75} & \underline{77.70} & \textbf{86.25} & \underline{73.29} & \textbf{72.95} & \underline{83.55} & \textbf{79.84} \\
    \bottomrule
    \end{tabular}
    \caption{
    ~\method performance on XLM-R-large in PAWS-X dataset under Limited-Resource Scenario. The metric in this table is accuracy. For each setting, we report the median scores among 5 runs.}
    \label{table:fewshot_results_pawsx}
\end{table*} 
\textbf{Ablation Study on Teacher Language Selection}

To evaluate the effectiveness of Teacher Language Selection, we conduct an ablation study using XLM-R-large as backbone. We reported average performance and cross-lingual transfer gaps of different language groups in Table~\ref{tab:ablation_of_teach}. It provides strong evidence for the effectiveness of our method.

Generally, the implementation of Teacher Language Selection in ALSACE significantly reduces the cross-lingual transfer gaps while improving performance across all languages, particularly for the student languages. 
It validates that, despite the efficacy of self-distillation, selecting adaptive teacher languages is crucial for boosting overall performance. With Teacher Language Selection, student languages achieve above-average improvements in both performance and cross-lingual transferability.

Specifically, when comparing ALSACE with other baselines, besides a performance improvement, there is a substantial reduction in the cross-lingual transfer gaps for all languages, particularly for student languages. 
ALSACE reduces the cross-lingual transfer gaps for student languages, ranging from 0.70 to 0.98 points and between 0.47 to 0.51 points for teacher languages.

Furthermore, excluding the teacher language selection diminishes the performance of student languages, limiting their ability to benefit from self-distillation. 
This results in an average performance decrease of 0.34 points and an increase of 0.73 points in cross-lingual transfer gaps for student languages.
ALSACE still outperforms the random selection by 0.42 points in performance and reduces cross-lingual transfer gaps for student languages by 0.77 points. These comparisons underscore the importance of selecting adaptive teacher languages.

Additionally, we remove some languages from distillation. 
First, we removed languages that exhibited weak performance from student languages. As expected, without the guidance of teacher languages, the performance of student languages remained poorly, with an observed increase in cross-lingual transfer gaps by 0.98 points. Subsequently, excluding languages with weak performance from teachers also led to a decrease in performance for both teachers and students by 0.34 and 0.63 points, respectively.
It underscores our hypothesis that the underperforming languages can serve as effective guidance for other languages due to closer linguistic proximity between languages.
Further details of the ablation study can be found in Appendix~\ref{appendix:ablation}.\\
\newline
\textbf{Ablation Study on Cross-lingual Self-Distillation}
To further investigate the source of performance increase and validate the effectiveness of the self-distillation, we conducted additional experiments with self-distillation methods as baselines with the following two settings:

\textbf{\textit{English-Only Self-Training}}~\cite{schick2020s}: We utilize the model fine-tuned on an English training set to produce pseudo-labels for the unlabeled English data used in cross-lingual self-distillation. Then, we choose the top 50\% of data with high confidence to fine-tune the model.

\textbf{\textit{Full-Language Self-Training}}: We generate pseudo-labels for translated multilingual data in all languages and select the top 50\% of multilingual data with high confidence to fine-tune the model.

We apply these two methods on mT5~\cite{xue2020mt5} and XLM-R ~\citep{conneau2019unsupervised} as baselines. As shown in Table~\ref{table:self-distillation-ablation}, \model outperforms all the self-distillation baselines on XNLI while improving the cross-lingual transferability of mPLMs, especially for the student languages.
It validates our method and indicates that ALSACE's improved performance stems from our self-distillation rather than from the incorporation of multilingual data. We also compared our method with other state-of-the-art self-distillation methods in Appendix~\ref{appendix:self-distillation}.



\subsection{Limited Resource Evaluation}
\begin{figure*}[hbpt]
    \centering
    \includegraphics[scale=0.138]{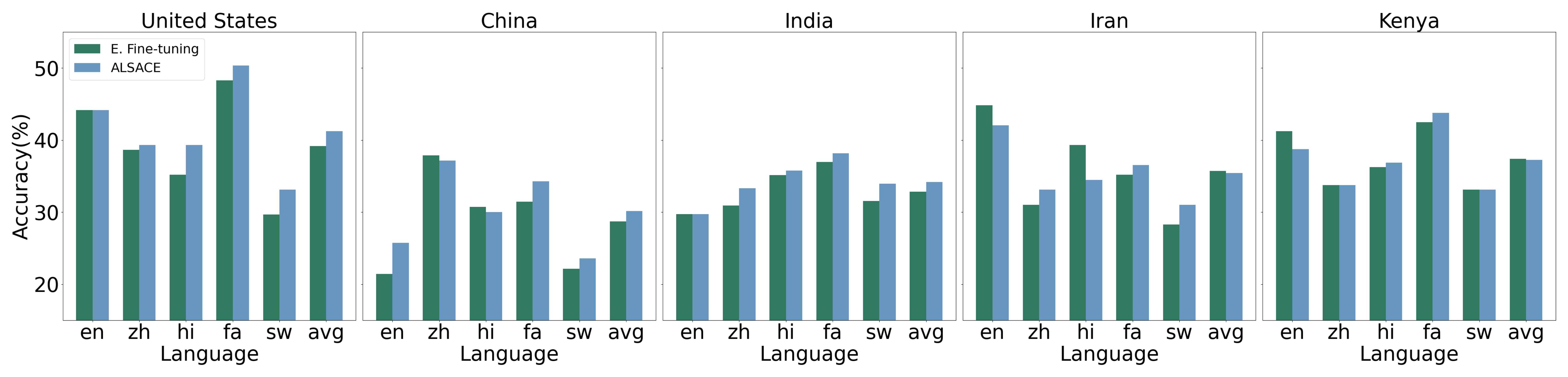}
    \caption{
Result of ~\method on XLM-R-large in GeoMLAMA dataset. 
The result shows that ~\method utilizes the teacher languages to guide other languages and generally improves their languages-specific knowledge.
    }\label{figure:geomlama_zlt_performance}
\end{figure*}
In scenarios with limited resources, where acquiring training data is extremely difficult (even for English), mitigating language-level performance disparities in mPLMs can be more challenging and crucial.
Therefore, to further evaluate the effectiveness of \model, we performed experiments on both XNLI and PAWS-X datasets in such scenarios.
Specifically, to simulate a limited resource scenario for XNLI, we fine-tune the mPLMs on $128$-shot English labeled examples as the baseline. Similarly, for PAWS-X, we fine-tune the mPLMs on $512$-shot English labeled examples. Further details can be found in Appendix~\ref{appendix:Details}.

To minimize the impact of the unlabeled multilingual parallel data used in ~\method, and thoroughly investigate the efficacy of self-distillation in~\method in limited resource situations, we also introduce two additional baselines: \textit{English-Only Self-Training}(E. Self-Train) and \textit{Full-Language Self-Training}(F. Self-Train). 
The results in Table~\ref{table:fewshot_results_xnli} and Table~\ref{table:fewshot_results_pawsx} despite that \model consistently improve the performance of all languages even when the training data is minimal. It underscores that \model improves model performance not by relying on the parallel corpora but by leveraging the knowledge of teacher languages gained from the mPLM pre-training stage, hence proving its robustness and efficiency in limited-resource settings.


\subsection{Analysis}
\label{subsection:analysis}


\begin{figure}
    \centering
    \includegraphics[scale=0.38]{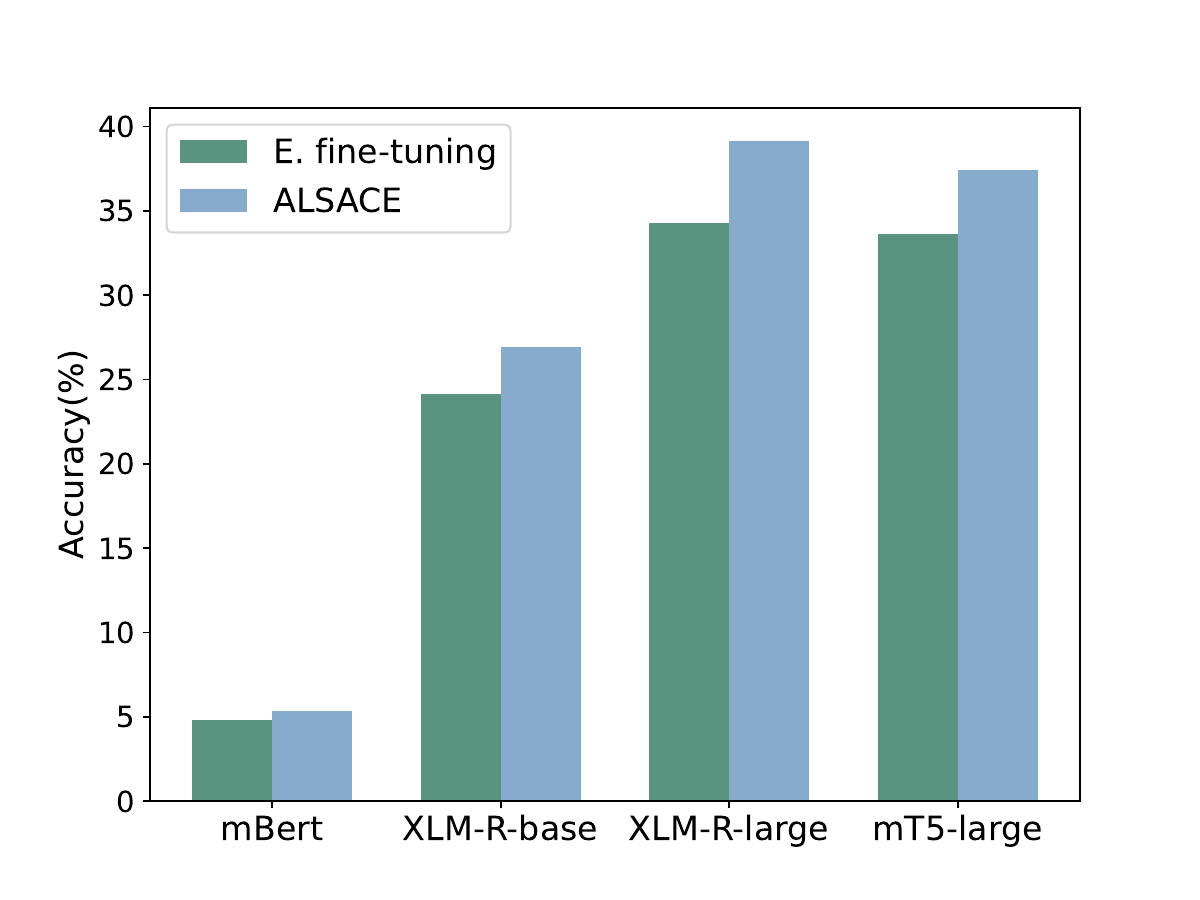}
    \caption{Accurately Answered Questions across All Languages in XNLI Baseline.}\label{fig:common_answer_question}
\end{figure}

\label{subsubsection:knowledge_probing}
The knowledge stored within mPLMs can be categorized into language-agnostic knowledge related to general tasks such as XNLI, which are based on logic and conceptual understanding, and language-specific knowledge related to specific linguistic and cultural factors.
In order to evaluate the \model's ability to alleviate performance disparity by reducing knowledge disparity and thereby improving overall performance, we conducted knowledge probing in GeoMLAMA to evaluate the changes in language-specific knowledge of mPLMs. We use the accuracy of question answers grouped according to countries and languages to measure the knowledge of mPLMs.

We examined the changes in language-specific knowledge gains before and after applying \method as shown in Figure~\ref{figure:geomlama_zlt_performance}. 
Results show that \model improves the performance of mPLM on knowledge probing tasks over various languages.
More details can be found in Table~\ref{table:Geomlama_table_result} in Appendix.

Notably, as shown in Figure~\ref{figure:method}, after applying Cross-lingual Self-Distillation, the specific knowledge of teacher languages can be transferred to other languages. It can be found out that under the guidance of teacher languages, other languages answer the geo-specific question correctly. For instance, as shown in the first sub-figure in Figure~\ref{figure:geomlama_zlt_performance}, English leverages its US-specific knowledge for other languages, leading to overall improvements for those respective languages. 
Similar results are observed in other sub-figures. This result strongly suggests that mPLMs capture far more knowledge than people previously believed, and language-specific knowledge remains a treasure for better alignment.

Furthermore, we explore whether \model successfully enhances language-agnostic knowledge over languages. Therefore, as demonstrated in Figure~\ref{fig:common_answer_question}, we evaluate the numbers of the accurately answered questions on the XNLI benchmark. This improvement demonstrates that the language-agnostic knowledge across different languages in mPLMs can mutually learn from each other. Our method reinforces the shared knowledge among the languages by bridging the knowledge disparity. As a result, we ensure that the efficacy of our method relies on alleviating the knowledge disparities across languages, including language-agnostic and language-special knowledge.
\section{Conclusion}
\label{sec:conclusion}
In this paper, we present \model, a simple yet effective method to address the language-level performance disparity in mPLMs.
\model mainly consists of two stages: Teacher Language Selection and Cross-Lingual Self-Distillation.
\model leverages the knowledge learned from the teacher languages to guide other languages and further improves the overall performance and cross-lingual transferability of mPLMs. 
Experiments show that ALSACE effectively mitigates language-level performance disparity and shows competitive performance on various multilingual datasets.
In addition, we further analyze each part of the \model to show the strengths of our proposed model.
Overall, \model is a promising approach to mitigating language-level performance disparity of mPLMs by utilizing self-distillation to reduce the performance disparity.

\section*{Limitation}
Our work has three limitations: 

1) We conduct experiments on a limited number of languages compared to the total number supported by mPLMs. Additionally, we only test other methods on the base and large model sizes of mT5 and XLM-R models. Therefore, in future work, we plan to extend our research to more languages and different mPLMs in different model sizes.

2) In the grand scheme of things, the languages we evaluate are relatively high-resource compared to some extremely low-resource languages such as Kaixana and Ainu. Improving our method on these extremely low-resource languages will be more exciting and meaningful. We plan to explore even more data-scarce settings in future work.

3) 
We use the cross-lingual transfer gap to measure mPLMs' cross-lingual transferability, aligning with prevailing research. However,if we reservedly enhances the performance of non-English languages while improving English greatly,
 the model's transfer gap could still be high despite the improvement in all languages.
Hence, we advocate for the development of the metric that can better reflect the performance equity and utility in multilingual models.

\section*{Acknowledgements}
We extend our heartfelt gratitude to the anonymous reviewers whose dedication and insightful feedback have significantly enhanced the caliber of this paper. Their constructive critiques and valuable suggestions were instrumental in refining our work. Additionally, we are deeply appreciative of the Program Chairs and Area Chairs for their meticulous handling of our submission and for their comprehensive and invaluable feedback. Their guidance has been pivotal in elevating the quality of our research.
This work is supported by the National Science Foundation of China under Grant No.61936012 and 61876004.

\bibliography{anthology,custom}


\appendix

\section{Appendix}
\label{sec:appendix}

\subsection{Experiment Details}
\label{appendix:Details}
\begin{table*}[h]
\centering
\resizebox{\textwidth}{!}{
\begin{tabular}{lcccccccccccccccc}
    \toprule
Method & en & fr & es & de & el & bg & ru & tr & ar & vi & th & zh & hi & sw & ur & avg \\
    \midrule
InfoXLM & \bf 0.80 & 0.90 & 0.70 & 1.00 & 1.00 & 0.50 & 0.60 & 1.70 & 1.40 & 0.40 & 1.10 & 1.10& 1.10 & 2.10 & 0.40 & 1.00 \\
ERNIE-M & 0.20 & 1.00 & 0.60 & 0.50 & 0.80 & 0.50 & 0.80 & 1.60 & 1.40 & 1.10 & 1.10 & 0.80 & 1.70 & 2.30 & 1.60 & 1.10 \\
ALSACE & 0.20 & \textbf{1.71} & \textbf{1.37} & .94 & \textbf{1.98} & \textbf{2.35} & \textbf{2.20} & \textbf{2.27} & \textbf{2.57} & \textbf{2.16} & \textbf{2.65} & \textbf{1.92} & \textbf{2.58} & \textbf{3.09} & \textbf{2.72} & \textbf{2.05} \\
    \bottomrule
\end{tabular}
}
\caption{Performance gain of each language compared with the initial XLM-R-large model.}
\label{table:method_scores}
\end{table*}

\begin{table*}[h]
\centering
\small
\resizebox{1\linewidth}{!}{\begin{tabular}{l c c c c c c c c c c c c c c c c}
\toprule
Model & en & fr & es & de & el & bg & ru & tr & ar & vi & th & zh & hi & sw & ur & avg \\
\midrule
PCT-XLM-R-large & 88.30 & 84.20 & 85.10 & 83.70 & 83.10 & 84.40 & 81.90 & 81.20 & 80.90 & 80.70 & 78.80 & 80.30 & 78.40 & 73.60 & 75.60 & 81.30 \\
\rowcolor{gray!15} $\Delta \downarrow$   & $\backslash$  & \multicolumn{1}{r}{4.10} & \multicolumn{1}{r}{3.20} & \multicolumn{1}{r}{4.60} & \multicolumn{1}{r}{5.20} & \multicolumn{1}{r}{3.90} & \multicolumn{1}{r}{6.40} & \multicolumn{1}{r}{7.10} & \multicolumn{1}{r}{7.40} & \multicolumn{1}{r}{7.60} & \multicolumn{1}{r}{9.50} & \multicolumn{1}{r}{8.00} & \multicolumn{1}{r}{9.90} & \multicolumn{1}{r}{14.70} & \multicolumn{1}{r}{12.70} & \multicolumn{1}{r}{7.45} \\\midrule
\model & 88.30 & 84.37 & 85.59 & 83.71 & 83.33 & 84.67 & 82.16 & 80.28 & 80.84 & 81.80 & 79.24 & 81.94 & 79.12 & 73.29 & 75.11 & \bf 81.58 \\
\rowcolor{gray!15} $\Delta  \downarrow$ & $\backslash$ & \multicolumn{1}{r}{3.93} & \multicolumn{1}{r}{2.71} & \multicolumn{1}{r}{4.59} & \multicolumn{1}{r}{4.97} & \multicolumn{1}{r}{3.63} & \multicolumn{1}{r}{6.15} & \multicolumn{1}{r}{8.02} & \multicolumn{1}{r}{7.46} & \multicolumn{1}{r}{6.51} & \multicolumn{1}{r}{9.06} & \multicolumn{1}{r}{6.37} & \multicolumn{1}{r}{9.18} & \multicolumn{1}{r}{15.01} & \multicolumn{1}{r}{13.19} & \multicolumn{1}{r}{\bf 7.20} \\
\bottomrule 
\end{tabular}}
\caption{Comparison of PCT-XLM-R-large and \model on XNLI benchmark across different languages. For a fair comparison, we report the performance of \model under the same setting of PCT. $\Delta$ represents the cross-lingual transfer gaps. A smaller gap indicates better cross-lingual transferability.}
\label{tab:comparison_pct_alsace}
\end{table*}
\textbf{Implement Details.} The unlabeled data used in \model is constructed by Supergen~\cite{meng2022generating}, which uses PLM to generate text guided by label-descriptive prompts. 
We use machine translation\footnote{The translation API from http://api.fanyi.baidu.com/ is used to generate the multilingual parallel data.} to generate unlabeled parallel multilingual text pairs based on the generated text. We leverage data generation methods(Supergen) and machine translation systems to construct an automatic pipeline for generating this valuable unlabeled parallel corpus at the lowest cost. 
We perform \model on mPLMs using $500$-shot unlabeled multilingual data with batch size $32$ on each language corresponding to the tasks of XNLI, PAWS-X, and XCOPA.
We set the learning rate to $3e-8$, and a dropout rate of $0.1$.
The thresholds $\theta$ in Equation \ref{eq:theta} are used to select the teacher languages are 0.06, 0.2, and 0.2 for XNLI, PAWS-X, and XCOPA, respectively. We set the threshold $\theta$ to be the average value of the language score $\hat{s_t}$ across all languages.

To evaluate the effectiveness of \model in limited resource scenarios,
we fine-tune the mPLMs for 100 epochs with learning-rate of $1e-6$ on $128$-shot English labeled examples as the baseline. 
Similarly, for PAWS-X, we fine-tune the mPLMs for 150 epochs with learning-rate of $1e-6$ on $512$-shot English labeled examples.

\subsection{Baselines}
\label{appendix:baselines}




\textbf{XLM-Align~\citep{chi2021improving}} presents denoising word alignment as a new cross-lingual pre-training task with 310M instances. It self-labels word alignments for parallel sentences and haphazardly masks tokens in a bitext pair for mPLMs to predict. 
\newline
\textbf{InfoXLM~\citep{chi2021infoxlm}} implements on the basis of mPLMs and tries to align the representation of multiple languages by introducing parallel corpora with a new pre-training task. Initializes its parameters with XLM-R and employs contrastive learning using 42GB parallel corpora to encourage encoded representations of bilingual sentence pairs to be more similar than negative examples.
\newline
\textbf{$\text{XLMR-adapter}_{256}$~\citep{he-etal-2021-effectiveness}} employs lightweight adapter modules on the XLM-R-large and achieves significant performances on low-resource and cross-lingual tasks.
\newline
\textbf{ERNIE-M~\citep{ouyang2021erniem}} is similar to InfoXLM and XLM-Align, which is implemented on the basis of XLM-R. It integrates back-translation into the pre-training process to encourage the model to align the representation of multiple languages with parallel corpora of about 68.8GB.
\newline
\textbf{VECO~\citep{luo2021veco}} plug a cross-attention module into the transformer encoder to explicitly build the interdependence between languages to pretrain a variable cross-lingual language model for both NLU and NLG.
\newline
\textbf{XLE~\citep{chi2022xlme}} use ELECTRA-style tasks for pre-training mPLMs with a generator and discriminator structure using 142B tokens. 





\begin{table*}[h]
\centering
\small
\resizebox{1\linewidth}{!}{\begin{tabular}{l l c c c c c c c c c c c c}
\toprule
Method & Setup & et & ht & id & it & qu & sw & ta & th & tr & vi & zh & avg \\
\midrule
XLM-R-base & CO-ZS & 54.40 & \bf49.80 & 56.00 & 54.80 & 49.00 & 53.40 & 51.40 & 57.40 & 55.00 & 54.20 & 57.40 & 53.89 \\
\model & CO-ZS & \bf 55.60 & 49.40 & \bf56.00 & \bf54.80 & \bf53.80 & \bf54.40 & \bf53.00 & \bf59.00 & \bf56.80 & \bf55.00 & \bf57.40 & \bf55.02 \\
\midrule
XLM-R-base & SI-CO-ZS & \bf61.40 & 51.60 & 66.60 & 64.40 & \bf49.60 & 55.80 & 62.00 & 61.60 & 60.20 &\bf 64.80 & 68.20 & 60.56 \\
\model & SI-CO-ZS & 58.40 & \bf53.40 & \bf66.40 & \bf65.80 & 49.00 & \bf57.40 & \bf62.60 & \bf62.60 & \bf62.40 & 64.60 & \bf 69.60 & \bf61.11 \\
\midrule
XLM-R-large & CO-ZS & 56.80 & (50) & 57.60 & \bf58.60 & (50) & \bf52.20 & 55.80 & 55.80 & 51.60 & 55.80 & 57.40 & 55.73 \\
\model & CO-ZS & \bf58.40 & (50) & \bf59.40 & 57.60 & (50) & 51.80 & \bf57.40 & \bf56.60 & \bf52.80 & \bf60.60 & \bf58.20 & \bf56.98 \\
\midrule
XLM-R-large & SI-CO-ZS & 72.00 & (50) & 77.00 & 77.20 & (50) & 61.60 & 67.20 & 76.40 & 74.40 & 76.60 & 77.40 & 73.31 \\
\model & SI-CO-ZS & \bf72.00 & (50) & \bf77.20 & \bf77.40 & (50) & \bf61.80 & \bf68.20 & \bf76.80 & \bf74.80 & \bf76.80 & \bf77.60 & \bf73.62 \\
\bottomrule
\end{tabular}}
\caption{Accuracy scores of different models on the XCOPA test set when transferring from English. Models are either only fine-tuned on the COPA~\citep{roemmele2011choice} training set and evaluated on different languages (CO-ZS) or fine-tuned first on SIQA~\citep{sap2019socialiqa} and then on COPA training set(SI-CO-ZS). Due to the inability of the XLM-R-large model to generate valid responses in Haitian Creole and Quechua, the scores for these languages are marked as (50) in the table.
}
\label{table:xcopa}
\end{table*}
\subsection{Compared with other state-of-art methods }
\subsubsection{Compared with pre-train based methods}
InfoXLM~\citep{chi2021infoxlm} initializes its parameters with XLM-R and employs contrastive learning using 42GB parallel corpora to encourage encoded representations of bilingual sentence pairs to be more similar than negative examples.

ERNIE-M~\citep{ouyang2021erniem} is implemented on the basis of XLM-R, and it integrates back-translation into the pre-training process to encourage the model to align the representation of multiple languages with parallel corpora of about 68.8GB.

While InfoXLM and ERNIE-M are built upon the basis of XLM-R by utilizing 42GB and 68.8GB data, respectively, our method only relies on a small amount of unlabeled parallel corpora (500-shot), which can be easily constructed with minimal effort.
Despite this minimal requirement, our approach achieves substantial enhancements compared to the baseline XLM-R model. Table~\ref{table:method_scores} illustrates the improvement of different methods across all languages on the XNLIdataset in comparison with the initial XLM-R-large baseline.

\subsubsection{Compared with self-distillation-based methods}
\label{appendix:self-distillation}
\citet{qi-etal-2022-enhancing} introduced PCT, a method that learns from various cross-lingual templates through a consistency loss, ensuring corresponding representations are aligned across languages. 
As indicated in Table~\ref{tab:comparison_pct_alsace}, our \model surpasses PCT-XLM-R-large in performance and demonstrates superior cross-lingual transferabilities. Thanks to the teacher language selection, \model not only minimizes the performance disparities among the student languages but also enables the teacher languages to benefit from self-distillation. This approach yields improved overall performance and narrows the cross-lingual transfer gaps more effectively than PCT-XLM-R-large.

\begin{figure}
    \centering
    \includegraphics[scale=0.33]{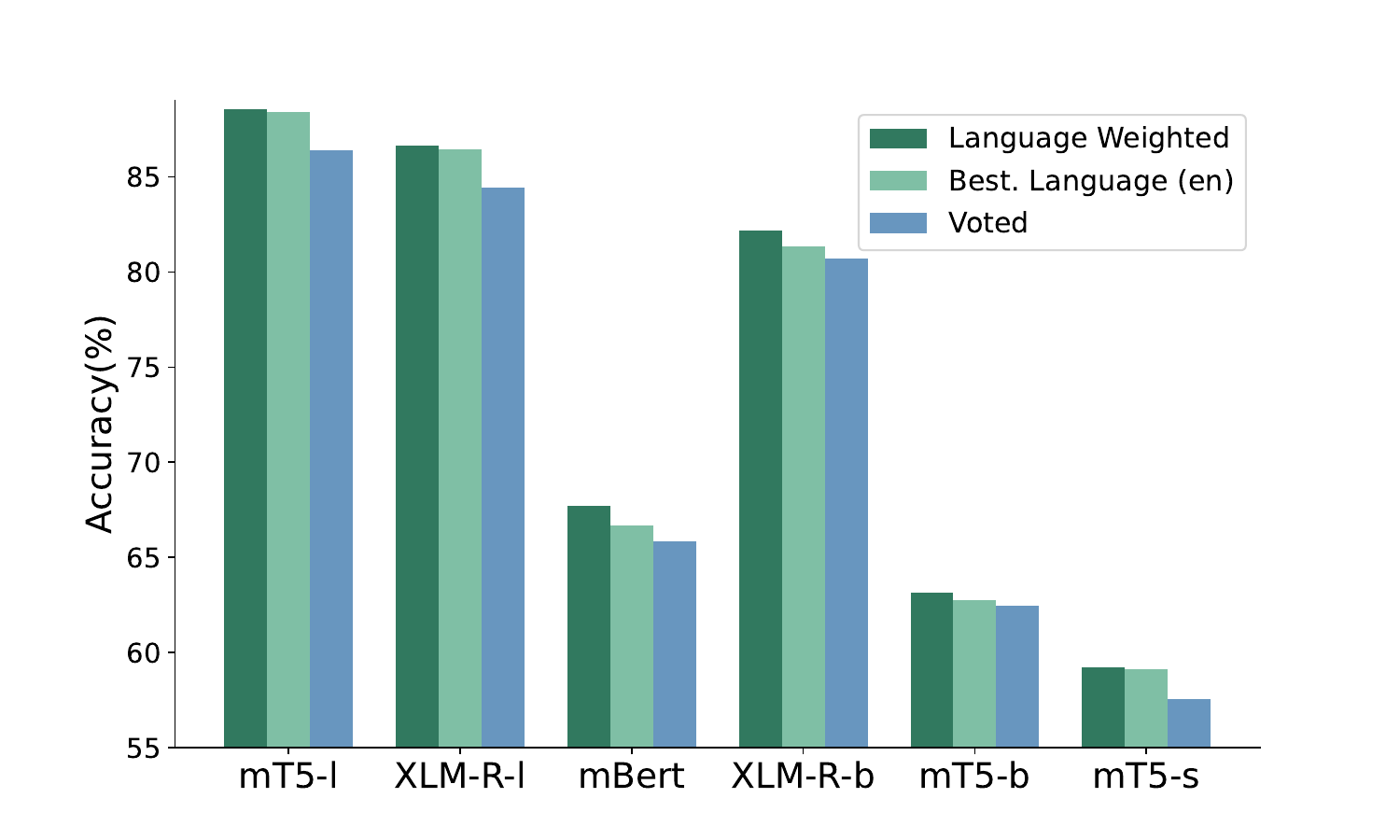}
    \caption{Performance of different Ensemble Methods.}\label{fig:ensemble_label}
\end{figure}
\subsection{Evaluation on XCOPA Benchmark}

We evaluate \model on the XCOPA benchmark, which is the causal commonsense reasoning benchmark across a range of typologically diverse languages, including both high and low-resource languages. 
Following the setting of \citet{ponti-etal-2020-xcopa}, models are either fine-tuned solely on the COPA~\citet{roemmele2011choice} training set and then evaluated on XCOPA's multilingual test sets or sequentially fine-tuned—initially on the SIQA dataset~\citet{sap2019socialiqa} followed by the COPA training set. Results in Table~\ref{table:xcopa} show that our method achieves substantial performance gains in most languages under various settings across different model sizes. These outcomes underscore the robustness and overall effectiveness of our method.

\begin{table*}[h!]
  \centering
\small
\resizebox{\textwidth}{!}{  
\begin{tabular}{lcccccccccccccccccc}
    \toprule
        {Method} & ~ & {en} & {fr} & {es} & {de} & {el} & {bg} & {ru} & {tr} & {ar} & {vi} & {th} & {zh} & {hi} & {sw} & {ur} & {avg.}\\
    \midrule
    \multirow{2}{*}{XLM-R-large} & Perf. & 86.45 & 80.89 & 81.83 & 81.22 & 79.36 & 80.74 & 78.78 & 77.22 & 77.44 & 77.41 & 75.53 & 77.82 & 74.56 & 69.62 & 70.86 & 77.98 \\
    & \graycell{} \graycell{$\Delta \downarrow$}  & \graycell{$\backslash$} & \graycell{5.56} & \graycell{4.62} & \graycell{5.23} & \graycell{7.09} & \graycell{5.71} & \graycell{7.67} & \graycell{9.23} & \graycell{9.01} & \graycell{9.05} & \graycell{10.93} & \graycell{8.63} & \graycell{11.90} & \graycell{16.83} & \graycell{15.60} & \graycell{8.47} \\

    \midrule
    \multirow{2}{*}{Excluding Weak Stu.} & Perf.& 86.79 & 81.70 & 83.19 & 82.32 & 80.86 & 82.55 & 80.54 & 79.08 & 78.36 & 79.22 & 77.07 & 79.12 & 76.45 & 71.62 & 72.71 & 79.44 \\
   & \graycell{$\Delta \downarrow$} & \graycell{$\backslash$} & \graycell{5.09} & \graycell{3.59} & \graycell{4.47} & \graycell{5.93} & \graycell{4.23} & \graycell{6.25} & \graycell{7.70} & \graycell{8.42} & \graycell{7.56} & \graycell{9.72} & \graycell{7.66} & \graycell{10.34} & \graycell{15.17} & \graycell{14.07} & \graycell{7.35} \\

    \midrule
    \multirow{2}{*}{Excluding Weak Tea.}&Perf. & 86.65 & 81.96 & 83.35 & 82.04 & 80.70 & 82.46 & 80.34 & 79.46 & 78.48 & 79.20 & 77.23 & 79.66 & 76.67 & 71.76 & 72.87 & 79.52 \\
    & \graycell{$\Delta \downarrow$} & \graycell{$\backslash$} & \graycell{4.69} & \graycell{3.29} & \graycell{4.61} & \graycell{5.95} & \graycell{4.19} & \graycell{6.31} & \graycell{7.19} & \graycell{8.16} & \graycell{7.45} & \graycell{9.42} & \graycell{6.99} & \graycell{9.98} & \graycell{14.89} & \graycell{13.77} & \graycell{7.13} \\

    \midrule
   \multirow{2}{*}{Random Selection} &Perf.& 86.81 & 82.15 & 83.15 & 82.18 & 80.94 & 82.66 & 80.78 & 79.40 & 79.14 & 79.08 & 77.78 & 79.34 & 75.93 & 71.89 & 72.91 & 79.61 \\
    & \graycell{$\Delta \downarrow$} & \graycell{$\backslash$} & \graycell{4.66} & \graycell{3.66} & \graycell{4.63} & \graycell{5.87} & \graycell{4.15} & \graycell{6.03} & \graycell{7.41} & \graycell{7.67} & \graycell{7.73} & \graycell{9.03} & \graycell{7.47} & \graycell{10.88} & \graycell{14.92} & \graycell{13.90} & \graycell{7.20} \\

    \midrule
   \multirow{2}{*}{No Selection} &  Perf. & 86.89 & 82.29 & 83.13 & 82.22 & 80.96 & 82.66 & 80.70 & 79.24 & 79.06 & 79.02 & 77.50 & 79.62 & 76.81 & 72.09 & 73.09 & 79.69 \\
   & \graycell{$\Delta \downarrow$} & \graycell{$\backslash$} & \graycell{4.60} & \graycell{3.76} & \graycell{4.67} & \graycell{5.93} & \graycell{4.23} & \graycell{6.19} & \graycell{7.65} & \graycell{7.83} & \graycell{7.87} & \graycell{9.39} & \graycell{7.27} & \graycell{10.08} & \graycell{14.80} & \graycell{13.80} & \graycell{7.20} \\

    \midrule
   \multirow{2}{*}{Scale-Based Selection} & Perf.& 86.93 & 82.38 & 83.01 & 82.24 & 81.02 & 82.73 & 80.84 & 79.32 & 78.80 & 79.46 & 77.62 & 79.70 & 76.97 & 71.94 & 73.23 & 79.75 \\
& \graycell{$\Delta \downarrow$} & \graycell{$\backslash$} & \graycell{4.55} & \graycell{3.91} & \graycell{4.69} & \graycell{5.91} & \graycell{4.19} & \graycell{6.09} & \graycell{7.60} & \graycell{8.12} & \graycell{7.47} & \graycell{9.30} & \graycell{7.23} & \graycell{9.96} & \graycell{14.99} & \graycell{13.69} & \graycell{7.18} \\

    \midrule
   \multirow{2}{*}{\model} & Perf.& 86.65 & 82.61 & 83.21 & 82.16 & 81.34 & 83.10 & 80.98 & 79.50 & 79.60 & 79.98 & 78.18 & 79.74 & 77.13 & 72.71 & 73.57 & \bf 80.03 \\
& \graycell{$\Delta \downarrow$} & \graycell{$\backslash$} & \graycell{4.04} & \graycell{3.44} & \graycell{4.49} & \graycell{5.31} & \graycell{3.55} & \graycell{5.67} & \graycell{7.15} & \graycell{7.05} & \graycell{6.67} & \graycell{8.47} & \graycell{6.91} & \graycell{9.52} & \graycell{13.94} & \graycell{13.08} & \graycell{\bf 6.62} \\

    \bottomrule
  \end{tabular}
  }
    \caption{Ablation Study of the Teacher Language Selection. $\Delta$ represents the cross-lingual transfer gaps, i.e., performance drop between English and other languages in zero-shot transfer. A smaller gap indicates better cross-lingual transferability. We report the average performance and cross-lingual transfer gaps of all languages.}
  \label{tab:tls_detail}
\end{table*}

\begin{table*}[h]
\centering
\small
\setlength\tabcolsep{3pt}
    \centering
    \resizebox{\linewidth}{!}{
        \begin{tabular}{lcccccccccccccccccc}
        \toprule
          Method & $\text{Params}$ & Perf. &  en & fr & es & de & el & bg & ru & tr & ar & vi & th & zh & hi & sw & ur & \textbf{avg($\Delta \downarrow$)} \\ 
        \midrule
          
            \multirow{2}{*}{XLM-R-base}& \multirow{2}{*}{225M}& ~   & 84.23 & 77.39 & 78.20 & 76.45 & 75.97 & 77.80 & 75.35 & 73.27 & 71.84 & 74.93 & 71.88 & 74.23 & 69.22 & 64.55 & 65.77 & 74.07 \\
~ & ~ & \graycell{$\Delta \downarrow$} & \graycell{$\backslash$} & \graycell{6.84} & \graycell{6.03} & \graycell{7.78} & \graycell{8.26} & \graycell{6.43} & \graycell{8.88} & \graycell{10.96} & \graycell{12.39} & \graycell{9.30} & \graycell{12.35} & \graycell{10.00} & \graycell{15.01} & \graycell{19.68} & \graycell{18.46} & \graycell{10.88 (\textcolor{red}{-})} \\

            \multirow{2}{*}{E. Self-Train.} &  \multirow{2}{*}{225M}& ~ &  84.09 & 77.96 & 78.28 & 76.73 & 76.25 & 78.14 & 75.65 & 73.33 & 72.12 & 75.27 & 71.78 & 74.35 & 69.54 & 64.85 & 66.27 & 74.31 \\
~ & ~& \graycell{$\Delta \downarrow$} & \graycell{$\backslash$} & \graycell{6.13} & \graycell{5.81} & \graycell{7.36} & \graycell{7.84} & \graycell{5.95} & \graycell{8.44} & \graycell{10.76} & \graycell{11.97} & \graycell{8.82} & \graycell{12.31} & \graycell{9.74} & \graycell{14.55} & \graycell{19.24} & \graycell{17.82} & \graycell{10.48 (\textcolor{green!70!black}{-0.40}))} \\
    \multirow{2}{*}{F. Self-Train.} &  \multirow{2}{*}{225M}& ~ &   84.13 & 78.18 & 78.40 & 76.85 & 76.51 & 78.16 & 75.67 & 73.81 & 72.04 & 75.33 & 71.84 & 74.57 & 69.74 & 64.91 & 66.57 & 74.45 \\

~ & ~ & \graycell{$\Delta \downarrow$} & \graycell{$\backslash$} & \graycell{5.95} & \graycell{5.73} & \graycell{7.28} & \graycell{7.62} & \graycell{5.97} & \graycell{8.46} & \graycell{10.32} & \graycell{12.09} & \graycell{8.80} & \graycell{12.29} & \graycell{9.56} & \graycell{14.39} & \graycell{19.22} & \graycell{17.56} & \graycell{10.37 (\textcolor{green!70!black}{-0.51})} \\

            \multirow{2}{*}{\method-base} &\multirow{2}{*}{225M}& ~  &  84.11 & 77.80 & 78.30 & 77.50 & 76.51 & 78.28 & 76.01 & 74.19 & 72.12 & 75.50 & 72.81 & 74.90 & 70.12 & 65.55 & 66.37 & 74.67 \\ 
    
~ & ~ & \graycell{$\Delta \downarrow$}  & \graycell{$\backslash$} & \graycell{6.31} & \graycell{5.81} & \graycell{\textbf{6.61}} & \graycell{\textbf{7.60}} & \graycell{\textbf{5.83}} & \graycell{\textbf{8.10}} & \graycell{\textbf{9.92}} & \graycell{\textbf{11.99}} & \graycell{\textbf{8.61}} & \graycell{\textbf{11.30}} & \graycell{\textbf{9.21}} & \graycell{\textbf{13.99}} & \graycell{\textbf{18.56}} & \graycell{\textbf{17.74}} & \graycell{\textbf{10.11 (\textcolor{green!70!black}{-0.77})}}  \\ 

            \midrule
            \multirow{2}{*}{XLM-R-large} &\multirow{2}{*}{550M}& ~  &  86.45 & 80.90 & 81.84 & 81.22 & 79.36 & 80.74 & 78.78 & 77.23 & 77.03 & 77.82 & 75.53 & 77.82 & 74.55 & 69.62 & 70.86 & 77.98 \\ 
    
~ & ~ & \graycell{$\Delta \downarrow$} & \graycell{$\backslash$} & \graycell{5.45} & \graycell{4.51} & \graycell{5.13} & \graycell{6.99} & \graycell{5.61} & \graycell{7.57} & \graycell{9.12} & \graycell{9.32} & \graycell{8.53} & \graycell{10.82} & \graycell{8.53} & \graycell{11.80} & \graycell{16.73} & \graycell{15.49} & \graycell{8.97 (\textcolor{red}{-})}\\ 

            \multirow{2}{*}{E. Self-Train.} &  \multirow{2}{*}{550M}& ~  & 86.77 & 81.44 & 82.32 & 81.40 & 79.92 & 81.16 & 79.18 & 78.10 & 77.54 & 78.42 & 76.19 & 78.46 & 75.31 & 70.48 & 72.04 & 78.58 \\
    
~ & ~ & \graycell{$\Delta \downarrow$} & \graycell{$\backslash$} & \graycell{5.33} & \graycell{4.45} & \graycell{5.37} & \graycell{6.85} & \graycell{5.61} & \graycell{7.59} & \graycell{8.67} & \graycell{9.23} & \graycell{8.35} & \graycell{10.58} & \graycell{8.31} & \graycell{11.46} & \graycell{16.29} & \graycell{14.73} & \graycell{8.77 (\textcolor{green!70!black}{-0.30})} \\

     \multirow{2}{*}{F. Self-Train.} &  \multirow{2}{*}{550M}& ~  & 86.81 & 81.54 & 82.69 & 81.68 & 80.30 & 81.96 & 79.70 & 78.40 & 78.12 & 78.90 & 76.45 & 78.06 & 74.93 & 70.34 & 71.90 & 78.79 \\
     
~ & ~ & \graycell{$\Delta\downarrow$} & \graycell{$\backslash$} & \graycell{5.27} & \graycell{4.12} & \graycell{5.13} & \graycell{6.51} & \graycell{4.85} & \graycell{7.11} & \graycell{8.41} & \graycell{8.69} & \graycell{7.91} & \graycell{10.36} & \graycell{8.75} & \graycell{11.88} & \graycell{16.47} & \graycell{14.91} & \graycell{8.60} (\textcolor{green!70!black}{-0.47}) \\

            \multirow{2}{*}{\method-large} &\multirow{2}{*}{550M}& ~  &  86.65 & 82.61 & 83.21 & 82.16 & 81.34 & 83.09 & 80.98 & 79.50 & 79.60 & 79.98 & 78.18 & 79.74 & 77.13 & 72.71 & 73.58 & 80.03 \\ 

~ & ~ & \graycell{$\Delta\downarrow$} & \graycell{$\backslash$} & \textbf{\graycell{4.04}} & \textbf{\graycell{3.44}} & \textbf{\graycell{4.49}} & \textbf{\graycell{5.31}} & \textbf{\graycell{3.56}} & \textbf{\graycell{5.67}} & \textbf{\graycell{7.15}} & \textbf{\graycell{7.05}} & \textbf{\graycell{6.67}} & \textbf{\graycell{8.47}} & \textbf{\graycell{6.91}} & \textbf{\graycell{9.52}} & \underline{\graycell{13.94}} & \textbf{\graycell{13.07}} & \textbf{\graycell{7.09}} (\textcolor{green!70!black}{-1.88}) \\ 

            \midrule
           \multirow{2}{*}{mT5-large}& \multirow{2}{*}{1.2B}& ~  &  88.42 & 82.44 & 83.49 & 81.68 & 81.14 & 81.96 & 79.90 & 77.33 & 76.87 & 78.52 & 75.31 & 77.74 & 75.31 & 72.63 & 70.88 & 78.91 \\

~ & ~ & \graycell{$\Delta\downarrow$} & \graycell{$\backslash$} & \graycell{5.98} & \graycell{4.93} & \graycell{6.74} & \graycell{7.28} & \graycell{6.46} & \graycell{8.52} & \graycell{11.09} & \graycell{11.55} & \graycell{9.90} & \graycell{13.11} & \graycell{10.68} & \graycell{13.11} & \graycell{15.79} & \graycell{17.54} & \graycell{10.19} (\textcolor{red}{-})\\ 

          \multirow{2}{*}{E. Self-Train.} &  \multirow{2}{*}{1.2B}& ~ & 88.50 & 82.46 & 84.33 & 82.02 & 81.84 & 82.34 & 80.96 & 78.04 & 78.06 & 79.70 & 76.81 & 78.44 & 75.73 & 73.57 & 72.04 & 79.66 \\
   
~ & ~ & \graycell{$\Delta\downarrow$} & \graycell{$\backslash$} & \graycell{6.04} & \graycell{4.17} & \graycell{6.48} & \graycell{6.66} & \graycell{6.16} & \graycell{7.54} & \graycell{10.46} & \graycell{10.44} & \graycell{8.80} & \graycell{11.69} & \graycell{10.06} & \graycell{12.77} & \graycell{14.93} & \graycell{16.46} & \graycell{9.48} (\textcolor{green!70!black}{-0.72}) \\
    \multirow{2}{*}{F. Self-Train.} &  \multirow{2}{*}{1.2B}& ~& 88.64 & 82.44 & 84.37 & 82.22 & 81.98 & 82.42 & 81.16 & 78.22 & 78.30 & 80.00 & 76.81 & 78.80 & 76.09 & 73.97 & 72.26 & 79.85 \\

    ~ & ~ & \graycell{$\Delta\downarrow$} & \graycell{$\backslash$} & \graycell{6.20} & \graycell{4.27} & \graycell{6.42} & \graycell{6.66} & \graycell{6.22} & \graycell{7.48} & \graycell{10.42} & \graycell{10.34} & \graycell{8.64} & \graycell{11.83} & \graycell{9.84} & \graycell{12.55} & \graycell{14.67} & \graycell{16.38} & \graycell{9.42} (\textcolor{green!70!black}{-0.77}) \\

            \multirow{2}{*}{\method-mT5} &\multirow{2}{*}{1.2B}& ~  &  88.60 & 83.69 & 84.79 & 83.17 & 82.91 & 83.91 & 81.80 & 79.54 & 78.84 & 80.20 & 77.90 & 80.92 & 77.25 & 75.17 & 73.13 & 80.79 \\ 

~ & ~ & \graycell{$\Delta \downarrow$} & \graycell{$\backslash$} & \underline{\graycell{4.91}} & \textbf{\graycell{3.81}} & \underline{\graycell{5.43}} & \textbf{\graycell{5.69}} & \textbf{\graycell{4.69}} & \textbf{\graycell{6.80}} & \underline{\graycell{9.06}} & \underline{\graycell{9.76}} & \underline{\graycell{8.40}} & \underline{\graycell{10.70}} & \textbf{\graycell{7.68}} & \textbf{\graycell{11.35}} & \textbf{\graycell{13.43}} & \textbf{\graycell{15.47}} & \textbf{\graycell{8.37}} (\textcolor{green!70!black}{-1.82}) \\ 
        \bottomrule
        \end{tabular}
    }
    \vspace{-0.1in}
    \caption{ Comparison of self-distillation baselines with \model. 
$\Delta$ represents the cross-lingual transfer gaps, i.e., performance drop between English and other languages in zero-shot transfer. A smaller gap indicates better cross-lingual transferability. We report the average performance and cross-lingual transfer gaps for all languages.}
    \label{table:self-distillation-ablation_whole}
\end{table*} 
\begin{table*}
\centering
\renewcommand{\arraystretch}{1.2}
\setlength\tabcolsep{3pt}
\small
\resizebox{1\linewidth}{!}{\begin{tabular}{ccccccccccccc}
\toprule
 Country& \multicolumn{2}{c}{US} & \multicolumn{2}{c}{CN}& \multicolumn{2}{c}{IN}& \multicolumn{2}{c}{IR}& \multicolumn{2}{c}{KE} &\multicolumn{2}{c}{Avg.}\\
\cmidrule(lr){2-3} \cmidrule(lr){4-5} \cmidrule(lr){6-7} \cmidrule(lr){8-9} \cmidrule(lr){10-11}\cmidrule(lr){12-13}
Method & XLM-R & \model & XLM-R & \model & XLM-R & \model & XLM-R & \model & XLM-R & \model & XLM-R & \model \\
\midrule
en & 0.4414 & \textbf{0.4414} & 0.2143 & \textbf{0.2571} & 0.2970 & \textbf{0.2970} & \textbf{0.4483} & 0.4207 & \textbf{0.4125} & 0.3875 & 0.3627 & \textbf{0.3607} \\
zh & 0.3862 & \textbf{0.3931} & \textbf{0.3786} & 0.3714 & 0.3091 & \textbf{0.3333} & 0.3103 & \textbf{0.3310} & 0.3375 & \textbf{0.3375} & 0.3443 & \textbf{0.3533}  \\
hi & 0.3517	& \textbf{0.3931} & \textbf{0.3071} & 0.3000 & 0.3515 & \textbf{0.3576} & \textbf{0.3931} & 0.3448 & 0.3625 & \textbf{0.3688} & \textbf{0.3532} & 0.3529 \\ 
fa & 0.4828	& \textbf{0.5034} & 0.3143 &\textbf{ 0.3429} & 0.3697 & \textbf{0.3818} & 0.3517 & \textbf{0.3655} & 0.4250 & \textbf{0.4375} & 0.3887 & \textbf{0.4062} \\ 
sw & 0.2966 & \textbf{0.3310} & 0.2214 & \textbf{0.2357} & 0.3152 & \textbf{0.3394} & 0.2828 & \textbf{0.3103} & 0.3313 & \textbf{0.3313} & 0.2894 & \textbf{0.3095} \\ 

\bottomrule

\end{tabular}
}
\caption{
Detailed results of ALSACE on XLM-R-large in GeoMLAMA dataset. The result shows that ALSACE utilizes teacher languages to guide others and generally improves their language-specific knowledge.
}
\label{table:Geomlama_table_result}
\end{table*}

\subsection{Analysis}
\label{appendix:analysis}
\paragraph{Why We Need to Select the Teacher Languages?}
To explore whether we need to select the teacher languages before transferring knowledge, we design an exploratory experiment on the XNLI dataset to demonstrate that selecting teacher language is necessary.
We measure the contribution of different ensemble strategies to model performance.
Specifically,
\textbf{\textit{language Weighted}}: For predicted labels and confidence scores from different languages, we use the confidence score of each language as weights and calculate the final ensemble prediction.

\textbf{\textit{Best Performing Language (en)}}: We use the results predicted by English as the final prediction.

\textbf{\textit{Voted}}: We give the same weight to the predicted labels for each language and get the final prediction result based on the voting result.

Figure~\ref{fig:ensemble_label} compares different multilingual models using different ensemble methods on the XNLI benchmark.
\textit{Voted} does not perform well due to noise from the under-performing student languages. On the other hand, by using the normalized language score $P(st)$ as weights for each language output in ensembling, it surpasses the performance of English, which is considered the best-performing high-resource language. 
This noteworthy discrepancy indicates that high-resource languages may not be suitable teacher languages. Besides high-resource languages, other languages also contribute to enhancing model performance.
Figure~\ref{figure:geomlama_zlt_performance} shows an experiment on GeomLAMA, demonstrating that high-resource languages may not be the most suitable for probing knowledge about a specific language condition. For instance, when addressing a query related to Chinese culture, Persian might yield a more accurate answer compared to English.


\begin{table}[h]
\centering
\small
\resizebox{0.9\linewidth}{!}{\begin{tabular}{cccc}
\toprule
\textbf{Lang.} & \textbf{Exc. Stu.(\%)} & \textbf{\model(\%)} & \textbf{Change($\Delta$)(\%)} \\
\midrule
en & 0.44 & 0.20 & -0.24 \\
fr & 1.40 & 1.72 & 0.32 \\
es & 1.30 & 1.38 & 0.08 \\
de & 1.00 & 0.94 & -0.06 \\
el & 1.60 & 1.98 & 0.38 \\
bg & 1.92 & 2.36 & 0.44 \\
ru & 1.92 & 2.20 & 0.28 \\
tr & 2.02 & 2.28 & 0.26 \\
zh & 1.80 & 1.92 & 0.12 \\
hi & 2.26 & 2.57 & 0.32 \\
vi & 1.62 & 2.16 & \textbf{0.54} \\
ar & 1.62 & 2.57 & \textbf{0.96 }\\
th & 1.98 & 2.65 & \textbf{0.68} \\
sw & 2.48 & 3.09 & \textbf{0.62} \\
ur & 2.24 & 2.71 & \textbf{0.48} \\
avg & 1.70 & 2.05 & 0.35 \\
\bottomrule
\end{tabular}
}
\caption{Ablation Study Comparison}
\label{tab:improvement}
\end{table}

\subsection{Ablation Study}
\label{appendix:ablation}
We conducted an ablation study to investigate the impact of teacher language selection, with detailed results provided in Figure~\ref{tab:tls_detail}. A comparison of ALSACE's performance with and without including student-student pairs indicates that even though there is a performance improvement when student-student pairs are excluded, a significant performance gap remains compared to the complete ALSACE model. This is particularly evident for student languages, as detailed in Table~\ref{tab:improvement}.
Additionally, when focusing on the student languages, such as Swahili and Urdu, the exclusion of student-student pairs results in comparatively diminished benefits from self-distillation.

The results clearly demonstrate that while the improvements persist, the performance of the ALSACE model employing randomly selected teacher languages still needs to catch up to the full ALSACE model across nearly all languages. This finding further underscores the efficacy of the teacher language selection strategy.
ALSACE demonstrates competitive performance across various baselines, achieving notable results even with a limited amount of unlabeled parallel data. We successfully alleviated the performance disparities among different languages.
As for the performance disparities, while there might still exist some gaps among different languages, ALSACE effectively mitigates these disparities, especially evident in languages like Swahili (sw), Urdu (ur), and Thai (th), as showcased in the performance comparison with English (en) in Table~\ref{table:main_results_xnli}. This aligns with our motivation to enhance cross-lingual transferability.

\begin{figure*}[hbpt]
    \centering
    \includegraphics[scale=0.31]{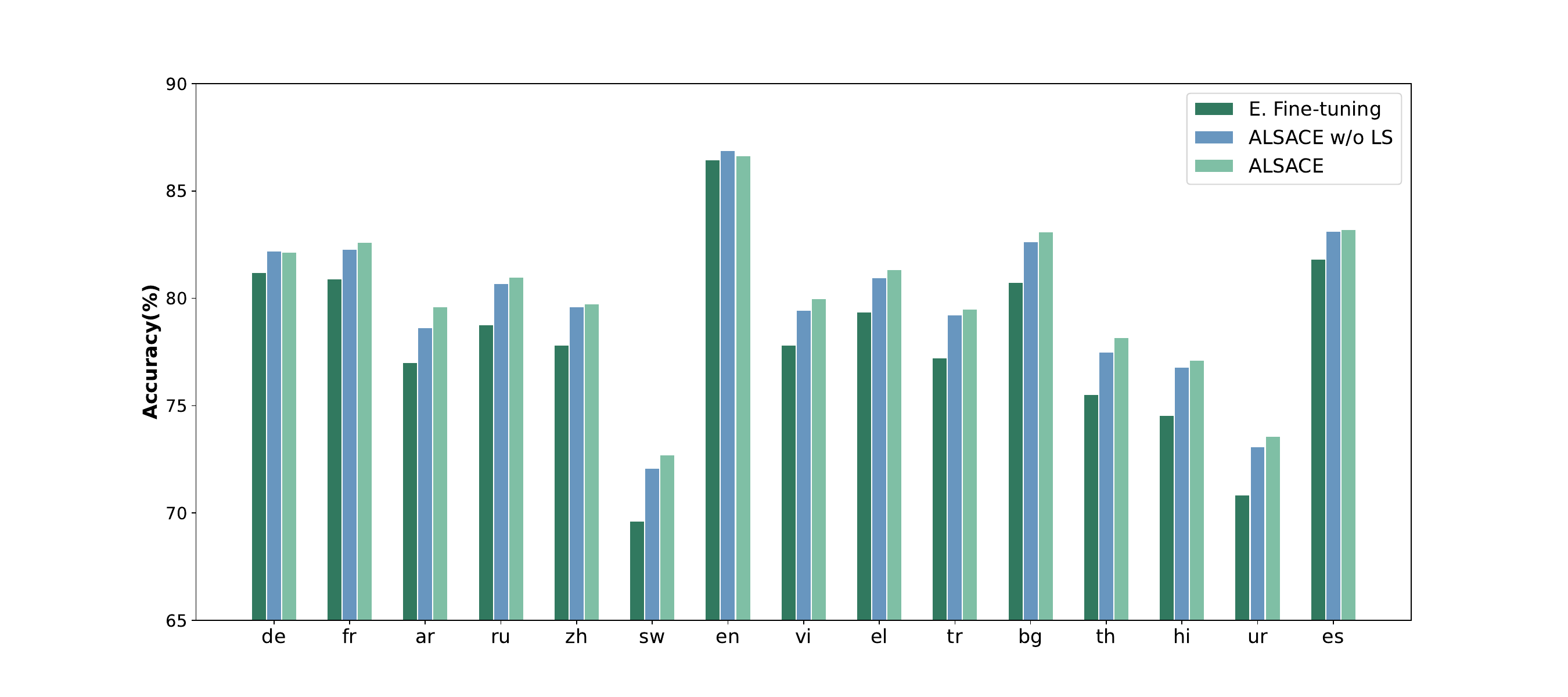}
    \caption{The Comparison of \model Performance with and without Language Selection on XNLI dataset set. All results are based on XLM-R-large.}\label{fig:language_selection_result}
\end{figure*}

\subsection{Geo-Diverse Commonsense across Countries}
Figure~\ref{figure:geomlama_zlt_performance} shows the detailed experiment results on GeomLAMA`~\citep{yin2022geomlama}, which demonstrates that \model improves the performance of mPLM on knowledge probing tasks over various languages.



\end{document}